\documentclass[10pt,twocolumn,letterpaper]{article}

\usepackage[pagenumbers]{iccv} 

\definecolor{iccvblue}{rgb}{0.21,0.49,0.74}
\usepackage[pagebackref,breaklinks,colorlinks,allcolors=iccvblue]{hyperref}

\def\paperID{17} 
\def\confName{ICCV}
\def\confYear{2025}

\title{PRISM: Product Retrieval In Shopping Carts using Hybrid Matching \vspace{-0.4cm}}
\author{Arda Kabadayi\\
Syracuse University\\
akabaday@syr.edu\\
\and
Jiajing Chen\\
Amazon\\
cjiajing@amazon.com
\and
Senem Velipasalar\\
Syracuse University\\
svelipas@syr.edu\\
}

\begin{document}
\maketitle
\begin{abstract}
Compared to traditional image retrieval tasks, product retrieval in retail settings is even more challenging.~Products of the same type from different brands may have highly similar visual appearances, and the query image may be taken from an angle that differs significantly from view angles of the stored catalog images.~Foundational models, such as CLIP and SigLIP, often struggle to distinguish these subtle but important local differences.~Pixel-wise matching methods, on the other hand, are computationally expensive and incur prohibitively high matching times. In this paper, we propose a new, hybrid method, called PRISM, for product retrieval in retail settings by leveraging the advantages of both vision-language model-based and pixel-wise matching approaches. To provide both efficiency/speed and fine-grained retrieval accuracy, PRISM consists of three stages: 
1) A vision-language model (SigLIP) is employed first to retrieve the top 35 most semantically similar products from a fixed gallery, thereby narrowing the search space significantly; 2) a segmentation model (YOLO-E) is applied to eliminate background clutter; 3) fine-grained pixel-level matching is performed using LightGlue across the filtered candidates.~This framework enables more accurate discrimination between products with high inter-class similarity by focusing on subtle visual cues often missed by global models. Experiments performed on the ABV dataset show that our proposed PRISM outperforms the state-of-the-art image retrieval methods by 4.21\% in top-1 accuracy while still remaining within the bounds of real-time processing for practical retail deployments.
\end{abstract}    
\section{Introduction} 
\label{sec:introduction}
    

    


\begin{figure}[t!]
    \centering
    \includegraphics[width=1.1\linewidth]{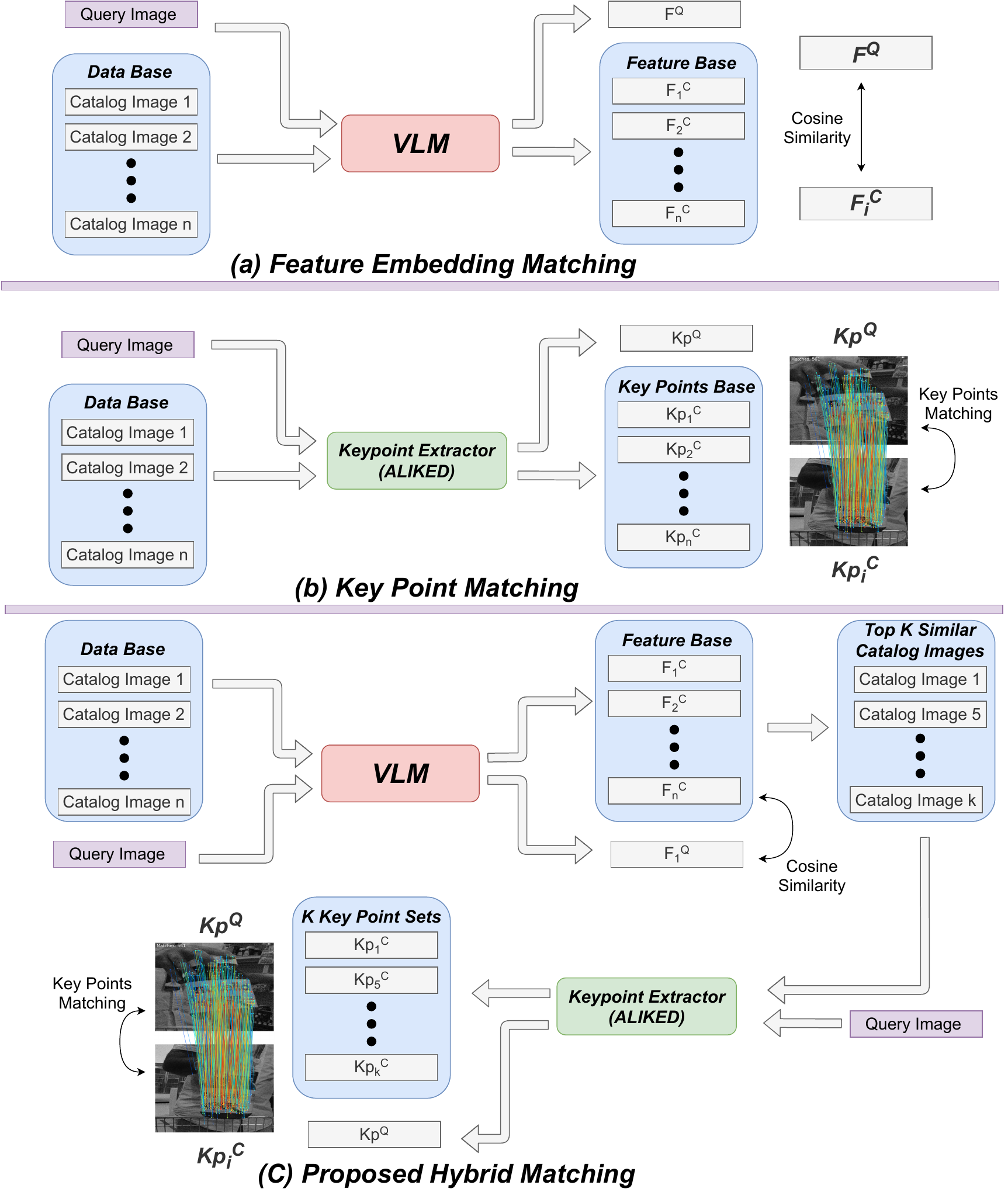}
    \vspace{-0.5cm}
    \caption{Comparison of retrieval approaches: (a) Matching of feature embeddings from a Vision-Language Model (VLM); (b) keypoint matching for finer-grain correspondence, (c) high-level overview of our hybrid PRISM. }
    \label{fig:best_code}
\vspace{-0.5cm}    
\end{figure}

With significant advancements in AI foundation models, such as CLIP~\cite{radford2021clip} and SigLIP~\cite{zhai2023sigmoid}, image matching and retrieval have achieved substantial progress recently. These vision-language models have demonstrated strong performance in extracting global semantic features from images, which allows them to find the top-k most semantically similar images in large galleries.
While these models are powerful and flexible, they struggle in specialized fields, such as medical imaging~\cite{zhao2023clip}, where the data differs significantly from natural images. Product retrieval in retail world is another application area of image matching/retrieval, but it presents the following additional challenges:
\textbf{(1) High inter-class similarity:} different products (e.g. from different brands) can have highly similar visual appearance. For example yogurt containers, cereal boxes, beverage bottles or soda cans often share similar shapes, colors, and packaging style; \textbf{(2) Varying image viewpoints:} the viewing angle of the query/customer image may differ significantly from that of the catalog/gallery images in the system; \textbf{(3) Domain gap:} captured query images from a certain sensor may differ from the catalog images, e.g. in terms of resolution and lighting; \textbf{(4) Occlusions:} customer/query images can be severely occluded blocking identifiable portions of the product. These factors introduce substantial difficulty in reliably retrieving the exact product from large catalogs, especially when subtle differences, such as shapes of the contents of a box (e.g. penne vs. fusilli pasta)  
are critical for identification. Since VLMs focus on global features, they often struggle to distinguish these subtle but important local differences. Thus, global vision-language models alone cannot fully overcome the aforementioned challenges.

Keypoint matching algorithms help address these challenges by enabling fine-grained, pixel-level comparisons that focus on detailed local features. However, their computational cost makes them impractical for directly comparing a query image against every image in very large product galleries. In addition, background feature points can distract and degrade the performance of keypoint matching without additional guidance.~Fig.~\ref{fig:best_code}(a) and Fig.~\ref{fig:best_code}(b) illustrate the difference between the VLM-based retrieval using image feature embeddings and fine-grained keypoint matching. VLM-based retrieval uses cosine similarity as the comparison metric, whereas keypoint matching employs a matching algorithm on the entire set of detected keypoints. 

Motivated by these and considering the strengths and weaknesses of the VLM- and keypoint-based methods, we propose a new, hybrid retrieval framework, referred to as the PRISM, for shopping cart-mounted camera systems in retail environments. Fig.~\ref{fig:best_code}(c) shows a high-level overview of PRISM, which takes advantage of both feature embedding matching and pixel-wise matching. The design of PRISM is based on the extensive experiments we performed on various VLM- and keypoint-based models. As will be discussed below, SigLIP is particularly good at choosing a subset of gallery images that are similar to the query, but struggle in correctly ranking them. Keypoint matching algorithms, on the other hand, are better at ranking. Equipped with these observations, PRISM is designed to have three-stages. In the first stage, SigLIP is used to extract semantic embeddings from the query image and retrieve the top-35 semantically most similar candidates from a product gallery. This step takes advantage of the strengths of SigLIP, efficiently narrows down the search space, and preserves high recall while significantly reducing the computational overhead. In the second stage, to mitigate the background noise potentially distracting keypoint matching, YOLO-E~\cite{wang2025yoloe} is employed to segment product regions in both query and candidate gallery images ensuring that comparisons focus only on the relevant object. Finally, the third stage utilizes LightGlue~\cite{lindenberger2023lightglue}, a robust local feature matcher, to perform detailed keypoint-level matching between the segmented product regions. This combination enables more accurate discrimination between products with high inter-class similarity by focusing on subtle visual cues often missed by global models. Thus, our proposed approach provides substantial improvement in top-1 accuracy of the fine-grained product retrieval in real-world retail scenarios. 

PRISM is designed with real-time performance in mind, making it suitable for deployment in fast-paced retail environments. Instead of relying only on global feature extractors or exhaustive pairwise matching, our method combines semantic filtering, product segmentation, and local feature matching for efficient and accurate retrieval. The contributions of this work include the following:
\begin{itemize}

\item We propose a new, three-step product retrieval system, PRISM, which combines the strengths of VLM- and keypoint matching-based approaches.~PRISM first uses a VLM to select a candidate list of similar gallery items, then separates the product region via segmentation, and finally applies detailed local feature matching to compare query and a subset of gallery images more precisely.

\item We present a system that balances retrieval accuracy with speed, making it practical for in-store applications.

\item We demonstrate the importance of segmentation masks in reducing background noise, thereby improving pixel-level keypoint matching accuracy.

\item We provide comprehensive experiments showing the role and contribution of each of the proposed three stages. 

\item We validate our approach on a real-world dataset, collected from shopping cart-mounted cameras, and show that it outperforms state-of-the-art VLM-based retrieval models, achieving significant gains in Top-1 accuracy.
\end{itemize}

Our experiments confirm that combining semantic-level retrieval with pixel-level local matching can bridge the gap between efficient search and accurate discrimination in fine-grained product recognition.

\section{Related Work}\label{sec:relatedwork}
\vspace{-0.1cm}
\subsection{Image Retrieval} \vspace{-0.05cm}
Early image retrieval methods relied on handcrafted local features.~SIFT~\cite{lowe2004distinctive} and SURF~\cite{park2022surf} capture distinctive keypoints for fine-grained matching. These approaches were effective but struggled with scalability and robustness to variations. With the rise of deep learning, Convolutional Neural Networks (CNNs) became popular for data-driven feature extraction, improving retrieval efficiency and robustness but often struggling to capture subtle visual differences between similar products \cite{babenko2014neural, tolias2015particular}. Vision-language models (VLMs), such as CLIP, further advanced retrieval performance by embedding images and text into a shared semantic space and enabling semantic search.~However, since they rely on global features, VLMs can struggle with fine-grained discrimination. On the other hand, local feature matching methods, such as SuperGlue \cite{sarlin2020superglue}, LoFTR, and LightGlue~\cite{sun2021loftr}, establish detailed pixel-level correspondences. They improve accuracy in fine-grained tasks at the cost of high computational complexity, which limits their scalability and real-time applicability.

\subsection{Vision-Language Models} \vspace{-0.1cm}
VLMs, such as CLIP, embed images and text into a common space, enriching image features with semantic context. Standard CLIP training uses a softmax-based contrastive loss. Recent works modify this loss to improve performance. SigLIP \cite{zhai2023sigmoid}, for instance, replaces the softmax with a pairwise sigmoid loss during CLIP pretraining. This change decouples the normalization across pairs, allowing much larger effective batch sizes and faster convergence. As a result, SigLIP achieves higher zero-shot image classification accuracy (e.g. 84.5\% top-1 on ImageNet) with modest compute.~Other CLIP-based methods explicitly target retrieval tasks by augmenting the embedding training.~JinaCLIP \cite{koukounas2025jina} uses multi-task contrastive training so that the same CLIP model excels at both image–text and text–text retrieval. This yields state-of-the-art (SOTA) performance on retrieval benchmarks, effectively turning CLIP into a text retriever without sacrificing its vision-language alignment.~LongCLIP~\cite{zhang2024longclip} adapts CLIP to much longer text inputs.~By stretching its positional embeddings and fine-tuning on ~1M long-caption image pairs, it can handle detailed prompts far beyond CLIP's 77 token limit. 
LongCLIP improves long text image retrieval by ~20\%.

\subsection{Image Segmentation} 
Earlier instance segmentation models combined object detectors with mask predictors.~Mask R-CNN~\cite{he2018maskrcnn} extends Faster R-CNN \cite{ren2016fasterrcnn}  by adding a parallel mask branch, enabling a detector to output both bounding boxes and pixel masks.~These methods require training in fixed object categories and often a separate detection and segmentation pipeline.~YOLO-E \cite{wang2025yoloe} (“Real-Time Seeing Anything”) is a recent YOLO-based model that simplifies this process by making segmentation promptable. It integrates an instance mask head into the YOLO architecture so that, given a text or image prompt, it produces both boxes and precise masks simultaneously. In effect, YOLO-E achieves open-vocabulary detection and segmentation in one efficient network, significantly improving earlier closed-set YOLO models. Segment Anything Model (SAM)~\cite{kirillov2023segment} is another segmentation foundation model trained on 11M images with 1.1B masks. SAM learns to segment any object given a user prompt (e.g., a point or bounding box) and exhibits strong zero-shot generalization, often matching or exceeding fully supervised baselines on new data.

\subsection{Pixel-wise Image Matching} \vspace{-0.1cm}
Recent matching approaches have moved beyond simple descriptor nearest-neighbor methods.~LoFTR \cite{sun2021loftr} is a transformer-based dense matcher that operates on coarsely sampled pixel grids. Instead of detecting keypoints, it computes feature descriptors conditioned on both images via self- and cross-attention, then extracts matches. This allows LoFTR to find correspondences even in low-texture or repetitive regions where traditional keypoint detectors fail. 
For sparse point matching, SuperGlue \cite{sarlin2020superglue} first applies a graph neural network to match sets of keypoint descriptors jointly, setting a new SOTA in local feature matching. LightGlue \cite{lindenberger2023lightglue} revisits SuperGlue’s design and introduces efficiency improvements: it dynamically stops the matching iterations on ``easy” image pairs and simplifies the network architecture. As a result, LightGlue runs much faster than SuperGlue while matching more keypoints. 
LightGlue's final optimized model achieves accuracy close to dense LoFTR at roughly 8× higher speed on outdoor images. This adaptive sparse matcher represents the best speed–accuracy trade-off among modern matching algorithms.

Yet, these pixel-wise image matching methods by themselves are not suitable for the problem of product retrieval/matching from shopping cart-mounted cameras in retail environments. Their computational cost makes them impractical for directly comparing a query image against every image in very large product galleries. Moreover, the challenges mentioned in Sec.~\ref{sec:introduction}, such as background features and noise, can degrade their performance. 
\section{Methodology} 
\label{sec:methodology}
\vspace{-0.1cm}
\begin{figure*}[t]
    \centering
    \includegraphics[width=\textwidth]{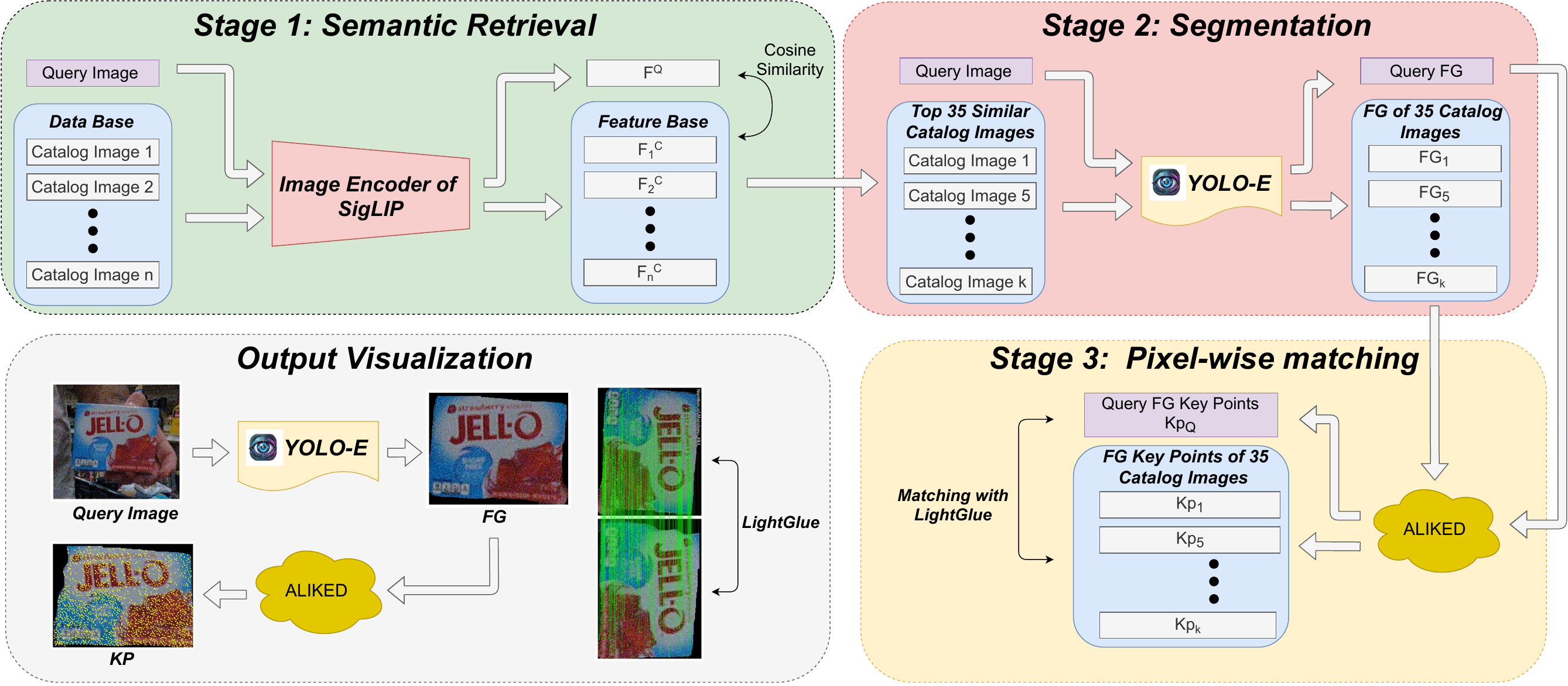}
    \caption{Overall architecture of the proposed PRISM. In stage 1, image feature embeddings $F$ are obtained by the image encoder of SigLIP. To narrow the search space, top 35 matches from the catalog are kept for further processing. In stage 2, YOLO-E is adopted to segment out foreground product region. In stage 3, LightGlue is used to establish pixel-level correspondences between the cropped query image and each of the cropped 35 candidate gallery images identified in Stage 1.
    }
    \label{fig:flow}
\end{figure*}

In order to address the challenges introduced by product retrieval from shopping cart-mounted camera systems, we propose a hybrid retrieval framework called PRISM. 
\subsection{Motivation} \vspace{-0.1cm}
We have performed initial experiments (summarized in first four rows of Tab.~\ref{tab:VLM_retrieval_results}) with various VLMs to compare their top-35, top-5, and top-1 accuracies. As can be seen, SigLIP achieves the best top-35 accuracy (0.744) among others, showing that it is better at choosing a good subset of gallery images that are similar to the query. However, as seen from the top-1 accuracy (0.386), SigLIP struggles with correctly ranking the returned images and choosing the best match. 

Pixel-wise matching algorithms, on the other hand, are better at ranking, but can be prohibitively slow since they compare a query image with every image in the gallery. For instance, LightGlue can take 1.35 minutes to match one product image in comparison to 21.7ms match time of SigLIP. 

Equipped with these observations, we propose PRISM as an hybrid framework to leverage the advantages of both VLM-based and pixel-wise matching approaches. To provide both efficiency/speed and fine-grained retrieval accuracy, PRISM consists of three stages, as shown in Fig.~\ref{fig:flow} and described below. 

\subsection{Stage 1: Semantic Retrieval} \vspace{-0.1cm}
In the first stage of our retrieval pipeline, we employ SigLIP to generate semantic embeddings from the query image and the gallery images. SigLIP is selected for this stage due to its superior performance in retrieval tasks, particularly in our product retrieval context. As shown in Table~\ref{tab:VLM_retrieval_results}, SigLIP achieves the highest Top-35 accuracy among several SOTA VLMs. By embedding images into a semantically rich and discriminative space, SigLIP helps ensure that the 35 retrieved candidates from the product gallery are visually and semantically  aligned with the query.

Let the query image be denoted as $q \in \mathbb{R}^{C \times W \times H}$, where $C$, $W$, and $H$ represent the number of channels, width, and height of the image, respectively (in our case, $C = 3$). The gallery consists of $N$ products, each represented by six different images taken from distinct angles, which are further discussed in Sec.~\ref{sec:experiments}. Hence, the complete gallery can be represented as $G \in \mathbb{R}^{(N \times 6) \times C \times W \times H}$. The SigLIP model maps an image $x \in \mathbb{R}^{C \times W \times H}$ into a normalized embedding vector $f(x) \in \mathbb{R}^d$, where $d$ is the embedding dimension. We apply this mapping to the query image and each of gallery images, resulting in a set of embeddings $f(q) \in \mathbb{R}^d, \,\, f(G) \in \mathbb{R}^{(N \times 6) \times d}$, respectively. 

To compute the similarity between the query and each gallery product image, we use the maximum cosine similarity between their embeddings. Specifically, for each gallery image embedding $f(G_i)$, we compute: \vspace{-0.2cm} 
\begin{equation} \small 
s_i = \text{cos-sim}\left(f(q), f(G_i)\right),
\end{equation} 
where $G_{i}$ is the $i^{th}$ gallery image. We then select the top 35 gallery items with the highest similarity scores ${s_i}$. This step takes advantage of the strengths of SigLIP, efficiently narrows down the search space for the subsequent matching stage, and preserves high recall while significantly reducing the computational overhead. More specifically, it reduces the candidate set from the full gallery with size $(N \times 6)$ (in our dataset $N = 394$) to a significantly smaller subset with size $(35 \times 6)$.

\subsection{Stage 2: Segmentation}
The second stage is aimed at addressing the challenge of noisy backgrounds and clutter typically found in retail images, where multiple products and shelf elements may appear together. To localize the product of interest within both the query image and candidate gallery images, and to mitigate the background noise potentially distracting the matching algorithm, we utilize the YOLO-E segmentation model, which 
outputs bounding boxes, segmentation masks and class labels. 

Let the input image be denoted as $I \in \mathbb{R}^{3 \times W \times H}$. The segmentation model \( S(\cdot) \) takes \( I \) and outputs a set of detected objects $
S(I) = \{ (b_k, m_k,  c_k ) \}_{k=1}^K$, 
where each detection \( k \) consists of:
\begin{itemize}
    \item \( b_k = (x_1, y_1, x_2, y_2) \) — the bounding box coordinates,
    \item \( m_k \in \{0,1\}^{W \times H} \) — the binary segmentation mask indicating the pixels belonging to the object,
    \item \( c_k \) — the predicted class label, which we do not use.
\end{itemize}

Since retail images can contain multiple products and complex backgrounds, selecting the correct object region is crucial. We therefore pick the detected bounding box \( b^* \) with the largest area, where $ b^* = \arg\max_{b_k} \text{Area}(b_k)$, 
and use the corresponding mask \( m^* \) to tightly crop the product region from the original image: \vspace{-0.2cm} 
\[
\small{
I^{\text{crop}} = I[y_1 : y_2, \, x_1 : x_2] \odot m^*,}
\]
where \( \odot \) denotes applying the mask to retain only the product pixels.

Empirically, as shown in Fig.~\ref{fig:segmentation-hist}, many false matches occur in background regions if the segmentation step is not applied, highlighting the importance of this stage.

\subsection{Stage 3: Pixel-wise matching}

In the third stage, we employ LightGlue, a fast, and accurate local feature matching algorithm, to establish pixel-level correspondences between the cropped query image and each of the cropped 35 candidate gallery images identified in Stage~1. This combination enables more accurate discrimination between products with high inter-class similarity by focusing on subtle visual cues often missed by global models. 

Let the cropped query image be denoted as $q^{crop} \in \mathbb{R}^{3 \times W^* \times H^*}$, 
and let the cropped gallery image be
\[
G^{crop}_i \in \mathbb{R}^{3 \times W_i^* \times H_i^*}, \quad \text{for } i = 1, \dots, 35.
\]
The LightGlue pipeline begins by extracting keypoints and descriptors from both images:
\[
\{ (k^q_j, d^q_j) \}_{j=1}^{N_q} = \phi(q^{crop}), \quad \{ (k^G_{i,l}, d^G_{i,l}) \}_{l=1}^{N_g} = \phi(G^{crop}_i),
\]
where:
\begin{itemize}
    \item \( k^q_j \in \mathbb{R}^2 \), \( k^G_{i,l} \in \mathbb{R}^2 \) are 2D keypoint locations,
    \item \( d^q_j, d^G_{i,l} \in \mathbb{R}^D \) are the corresponding feature descriptors,
    \item \( \phi(\cdot) \) represents the keypoint extractor (ALIKED \cite{zhao2023aliked} in our case),
    \item \( N_q \) and \( N_g \) are the number of detected keypoints in the query and gallery images, respectively.
\end{itemize}

LightGlue performs a sparse matching operation to generate a set of candidate correspondences:
\[
M_i = \{ (k^q_j, k^g_{i,l}) \}_{j,l} \quad \text{such that } d^q_j \sim d^G_{i,l}.
\]

To further filter out inconsistent or noisy matches, we apply RANSAC and retain only the inlier matches:
\[
M^{\text{inliers}}_i = \text{RANSAC}(M_i).
\]

Finally, we select the gallery image \( G_{i^*} \) with the highest number of inlier matches:
\[
i^* = \arg\max_i \left| M^{\text{inliers}}_i \right|.
\]

As a result, using LightGlue enables us to accurately differentiate between highly similar products, particularly when subtle variations (e.g., a logo change or flavor label) are the only discriminative cues. This step significantly reduces false matches and improves the robustness of our retrieval pipeline under challenging conditions such as occlusions, varying viewpoints, and lighting inconsistencies.

\section{Experiments}
\label{sec:experiments}
We conduct our product retrieval experiments on the ABV dataset \footnote{https://physicalstoreworkshop.github.io/challenge.html}, which contains images of various products taken directly in real store environments. These images often include multiple items and cluttered backgrounds, making the retrieval task more challenging. Moreover, the dataset reflects real-world conditions such as occlusions, viewpoint variations, and inconsistent lighting, providing a robust benchmark for evaluating retrieval performance. The dataset includes product photos captured from different angles, specifically \text{back\_drop}, \text{bottom\_drop}, \text{front\_drop}, \text{front\_view}, \text{side\_drop}, \text{and top\_drop}, where \_drop refers to someone dropping an item to cart, and \_view refers to an item directly being shown to the camera. The number of images per product varies, but most products have at least 50 total images and at least one image per defined angle. For each product, we use one image per different view angle, and form a six-image gallery set per product to construct a comprehensive multi-view representation. The remaining images are treated as separate customer queries, simulating real-world scenarios in which customers may submit photos taken from arbitrary angles. 

In our experiments, we compare various retrieval models and configurations to evaluate the performance of our proposed product matching system. The goal is to determine how well different vision-language models and pixel-level matching algorithms can retrieve the correct product from a gallery, given a customer query image. We first perform a comparison of various VLMs, namely CLIP, SigLIP, JinaCLIP and LongCLIP, in terms of Top-35, Top-5, and Top-1 accuracy metrics, which respectively measure whether the correct item appears in the top 35, 5, or 1 retrieved results, as well as the average match time per query image (measured using an NVIDIA A100 GPU).

As shown in Table~\ref{tab:VLM_retrieval_results}, CLIP struggles with fine-grained visual distinctions between products, achieving only 0.5613 Top-35, 0.3137 Top-5, and 0.1704 Top-1 accuracy. 
SigLIP significantly outperforms CLIP in our experiments, achieving 0.7442, 0.5271 and 0.3858 in Top-35, Top-5 and Top-1 accuracy, respectively. Two more recent models, namely JinaCLIP and LongCLIP, extend the CLIP architecture with various optimizations. JinaCLIP improves upon CLIP’s alignment of text and image embeddings by modifying its training pipeline and data, whereas LongCLIP adapts CLIP to much longer text inputs. 
Yet, these models do not work well on the product retrieval from shopping cart cameras. JinaCLIP provides 0.5304 Top-35 and 0.2254 Top-1 accuracy. While LongCLIP performs better than JinaCLIP with 0.7292 Top-35 and 0.3675 Top-1 accuracy, it is outperformed by SigLIP. 
\begin{table}[h!] \vspace{-0.22cm}
\centering
\resizebox{\columnwidth}{!}{%
\begin{tabular}{|l|c|c|c|c|}
\hline
\textbf{Method} & \textbf{Top-35 Acc} & \textbf{Top-5 Acc} & \textbf{Top-1 Acc} & \textbf{Speed} \\
\hline
CLIP \cite{radford2021clip} & 0.5613 & 0.3137 & 0.1704 & 25ms \\
SigLIP \cite{zhai2023sigmoid} & 0.7442 & \textbf{0.5271} & 0.3858 & 21.7ms \\
JinaCLIP \cite{koukounas2025jina} & 0.5304 & 0.3279 & 0.2254 & 55.4ms \\
LongCLIP \cite{zhang2024longclip} & 0.7292 & 0.5242 & 0.3675 & \textbf{19.6ms} \\
PRISM (Ours) & 0.7442 & 0.5142 & \textbf{0.4279} & 725ms \\
\hline
\end{tabular}
}
\vspace{-0.3cm}
\caption{Comparison of retrieval accuracy (Top-35, Top-5, and Top-1) and average inference time per query.}
\label{tab:VLM_retrieval_results}
\vspace{-0.25cm}
\end{table}

The last row of Tab.~\ref{tab:VLM_retrieval_results} shows the performance of our proposed PRISM. As can be seen, by leveraging the strengths of SigLIP and combining it with the better ranking ability of LightGlue, our approach provides the best Top-1 accuracy of 0.4279, outperforming the closest baseline SigLIP by 4.21\%. As will be shown in the Ablation Studies, incorporating the segmentation in the second stage helps filter irrelevant background features and improves fine-grained pixel-level matching.~Our approach provides strong retrieval performance, especially when precise localization of a customer’s product is critical. Although our method requires longer matching time (725ms per query), it is much faster than only using a keypoint matching-based approach (which can take about 1.3 minutes per query), and still remains within the bounds of real-time processing for practical retail deployments, where accurate product retrieval is often more critical than marginal differences in latency. 

\begin{figure}[t!]
\centering
\setlength{\tabcolsep}{0.2pt}
\renewcommand{\arraystretch}{1.0}
\begin{tabular}{ccc}
\textbf{Customer } & \textbf{SIGLIP } & \textbf{Our Method } \\
\textbf{ Image} & \textbf{ (Incorrect)} & \textbf{ (Correct)} \\
\includegraphics[width=5cm,height=3.2cm,keepaspectratio]{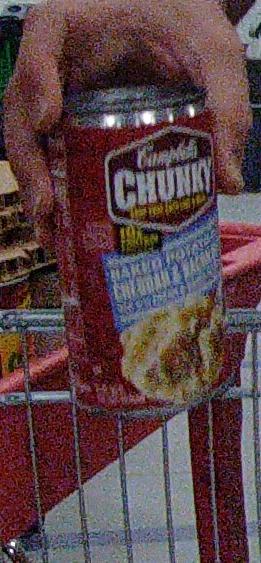} &
\includegraphics[width=5cm,height=3.2cm,keepaspectratio]{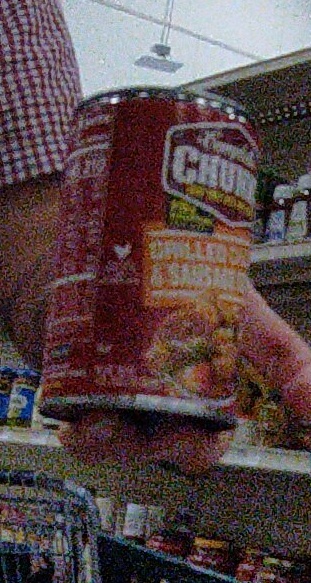} &
\includegraphics[width=5cm,height=3.2cm,keepaspectratio]{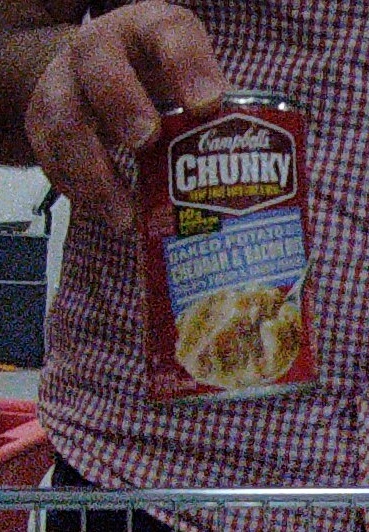} \\

\includegraphics[width=5cm,height=3.4cm,keepaspectratio]{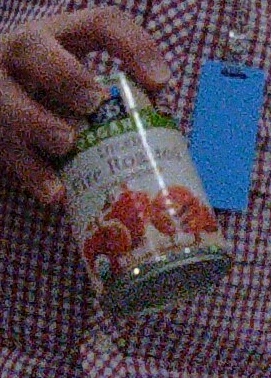} &
\includegraphics[width=5cm,height=3.4cm,keepaspectratio]{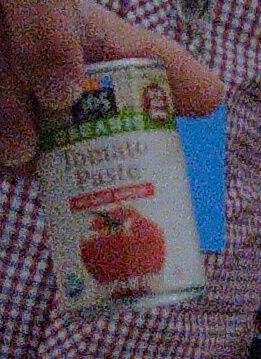} &
\includegraphics[width=5cm,height=3.4cm,keepaspectratio]{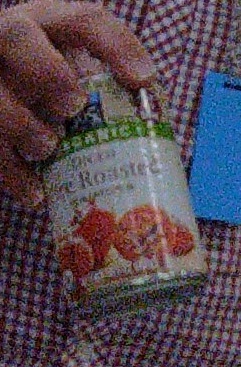} \\

\includegraphics[width=5cm,height=3.5cm,keepaspectratio]{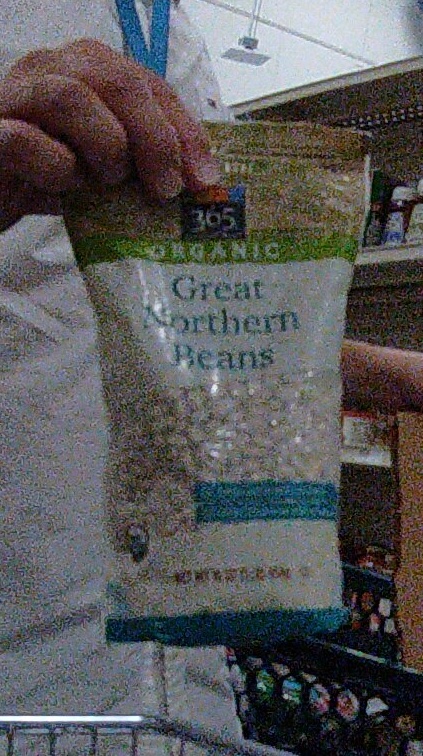} &
\includegraphics[width=5cm,height=3.5cm,keepaspectratio]{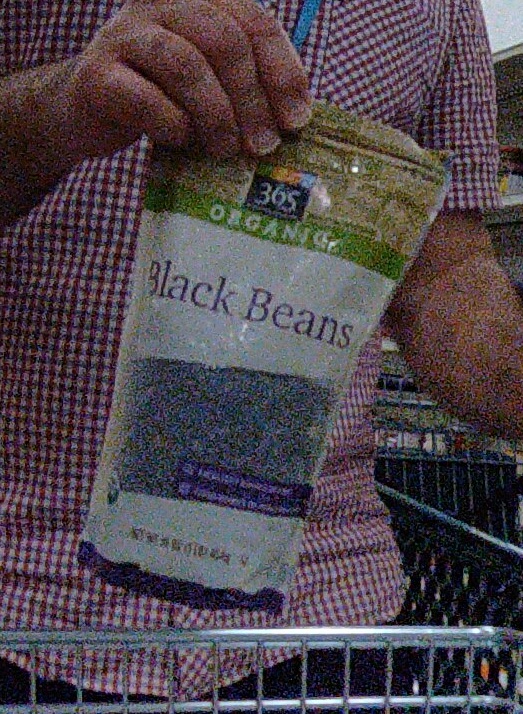} &
\includegraphics[width=5cm,height=3.5cm,keepaspectratio]{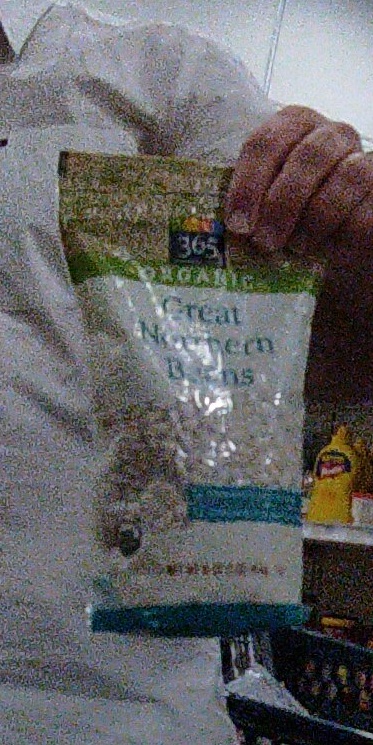} \\

\includegraphics[width=5cm,height=2.4cm,keepaspectratio]{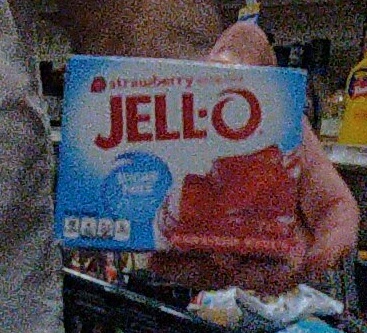} &
\includegraphics[width=5cm,height=2.4cm,keepaspectratio]{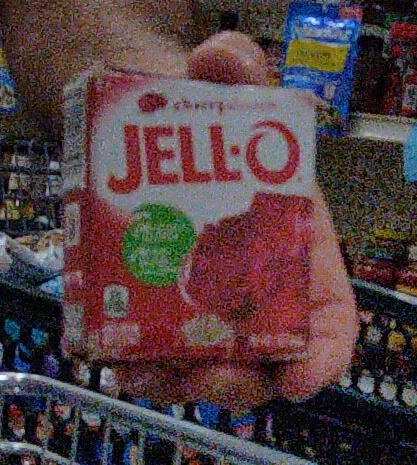} &
\includegraphics[width=5cm,height=2.4cm,keepaspectratio]{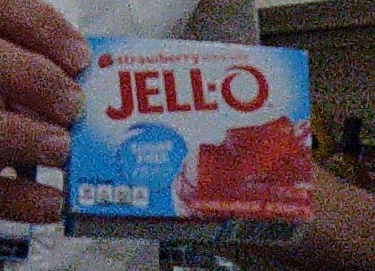} \\

\includegraphics[width=5cm,height=3.5cm,keepaspectratio]{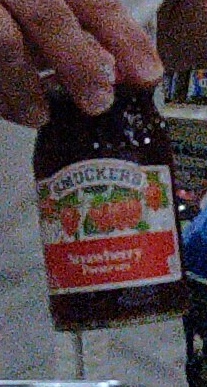} &
\includegraphics[width=4cm,height=3.5cm,keepaspectratio]{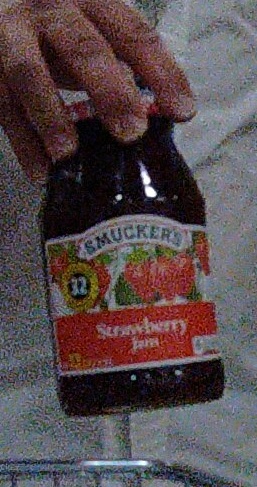} &
\includegraphics[width=4cm,height=3.5cm,keepaspectratio]{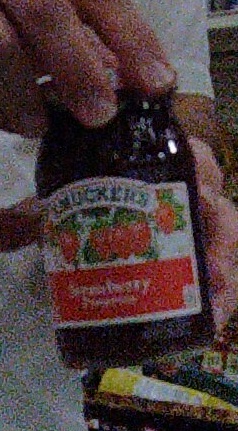} \\
\end{tabular}
\vspace{-0.4cm}
\caption{Qualitative results comparing our approach with the second best performer SigLIP. From left to right: customer query image, top-1 gallery image retrieved by SigLIP (incorrect), and top-1 gallery image retrieved by our method (correct).}
\label{fig:qualitative_comparison}
\vspace{-0.4cm}
\end{figure}

\begin{figure}[htb]
\centering
\setlength{\tabcolsep}{0.2pt}
\renewcommand{\arraystretch}{1.0}

\begin{subfigure}{\linewidth}
    \centering
    \begin{tabular}{cc}
        \includegraphics[width=4.2cm, height=4cm]{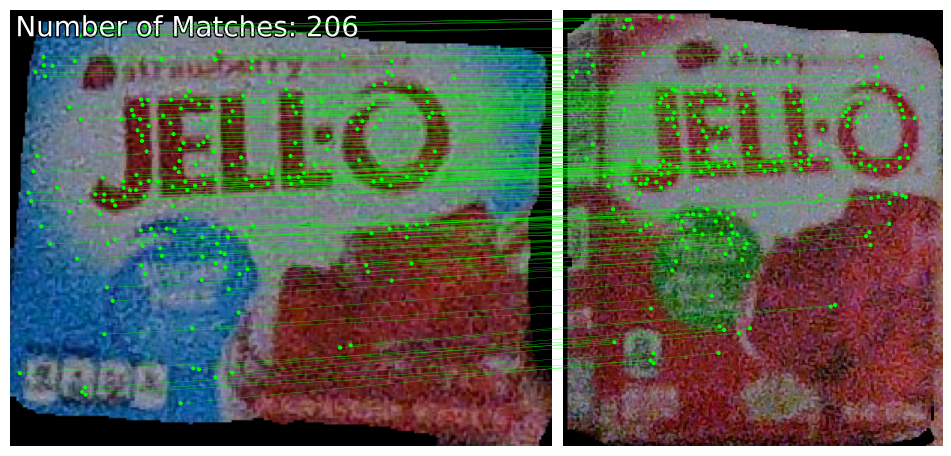} &
        \includegraphics[width=4.2cm, height=4cm]{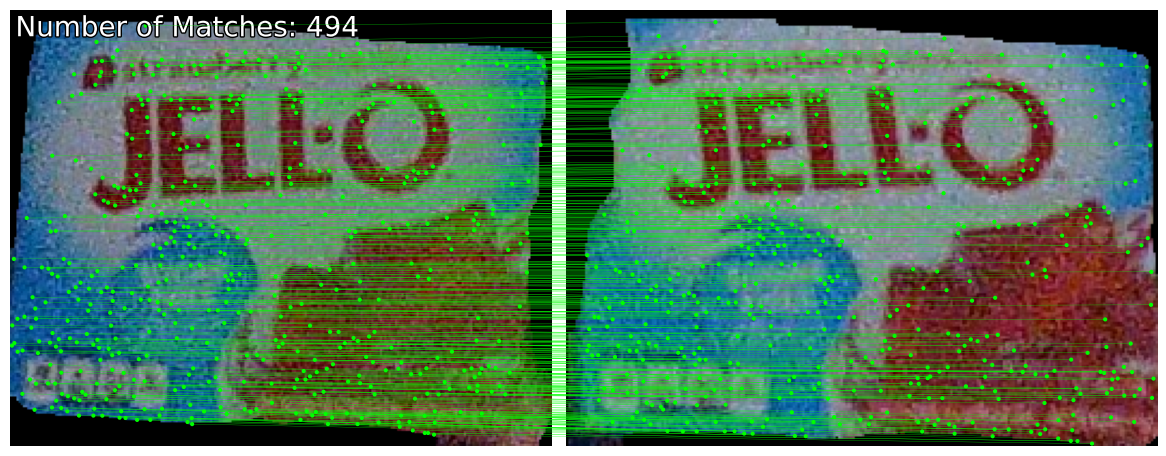}
    \end{tabular} \vspace{-0.2cm}
    \caption{First two images show the LightGlue matches across the wrong pair - only 206 matches. 2nd image is the SigLIP's top-1 return if not followed by LightGlue. Last two images show the correct query-gallery match with 494 matched points. 4th image is the PRISM's top-1 return.}
\end{subfigure}

\vspace{0.3em} 

\begin{subfigure}{\linewidth}
    \centering
    \begin{tabular}{cc}
        \includegraphics[width=4.2cm, height=4cm]{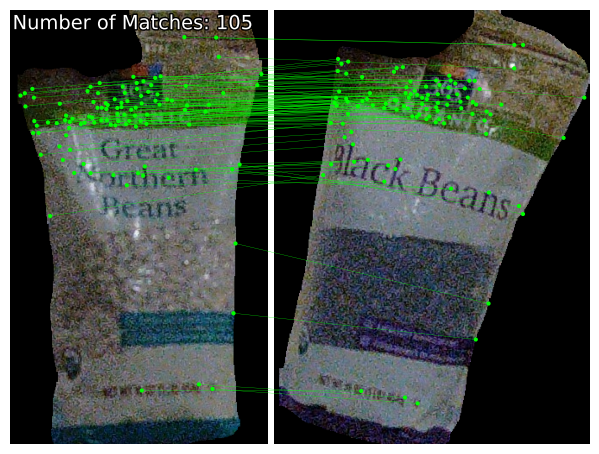} &
        \includegraphics[width=4.2cm, height=4cm]{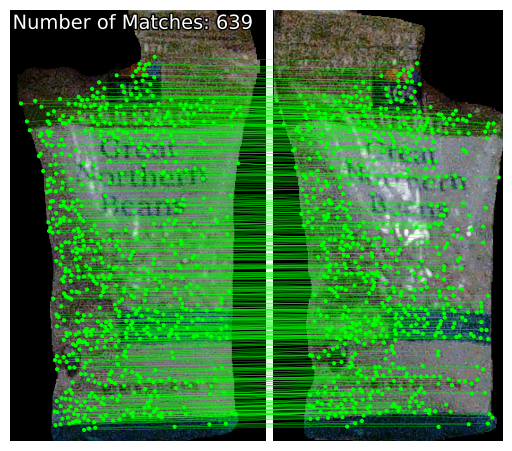}
    \end{tabular} \vspace{-0.2cm}
    \caption{First two images show the LightGlue matches across the wrong pair- only 105 matches. 2nd image is the SigLIP's top-1 return if not followed by LightGlue. Last two images show the correct query-gallery match with 639 matched points. 4th image is the PRISM's top-1 return.}
\end{subfigure}
\vspace{-0.7cm}
\caption{Comparison of the LightGlue matches for PRISM's top-1 return and for SigLIP's top-1 return.}
\label{fig:LightGlue_comparison}
\vspace{-0.3cm}
\end{figure}

Providing highest Top-1 accuracy
is a crucial advantage in practical retail environments, where only the top prediction is used for downstream tasks, such as automated checkout or inventory tracking. To better understand this performance difference, we conducted a qualitative analysis focusing on the subset of products that SigLIP retrieves correctly within its Top-35 predictions, but fails to rank as the top result, while our model retrieves them correctly at the Top-1 position. As illustrated in Fig.~\ref{fig:qualitative_comparison}, SigLIP often selects products from the same brand that look nearly identical but differ in fine-grained attributes, such as flavor, type, text, or packaging size (e.g. great northern beans vs. black beans as shown in the 3rd row of Fig.~\ref{fig:qualitative_comparison}). These subtle variations are often critical in retail settings, yet they are overlooked by global feature-based models. In contrast, our proposed PRISM effectively captures these small differences. This enables our model to distinguish between similar-looking products and identify the correct one with higher precision, especially in visually dense or cluttered environments. Fig.~\ref{fig:LightGlue_comparison} shows a comparison of the matched points between the query image and PRISM’s top-1 return and SigLIP’s top-1 return, showing that applying LightGlue on the top 35 retrieved images allows capturing subtle differences.


\section{Ablation Studies}
\label{sec:ablationstudies} \vspace{-0.1cm}

To better understand the contribution of each component in our pipeline, we conduct a series of ablation studies. These include comparisons between different segmentation models, matching algorithms, and preprocessing strategies. Each study isolates a specific design choice and analyzes its impact on the retrieval performance. 

\subsection{Importance of Stage 1 - Semantic Retrieval}
To further understand the contribution of the semantic retrieval/filtering (Stage 1), we compare our full pipeline, adopting SigLIP, YOLO-E and LightGlue, with a variant that removes the vision-language model SigLIP and instead relies solely on YOLO-E for class-based narrowing of the gallery. Specifically, we first finetune YOLO-E using our dataset with three manually-labeled high-level product classes: \{\textit{bagged}, \textit{bottled}, \textit{canned}\}. Then, we use YOLO-E to predict the product category (bagged, bottled, or canned) and restrict the gallery to images belonging to that category before applying LightGlue-based matching.
\begin{table}[ht!]
\centering
\resizebox{\columnwidth}{!}{%
\begin{tabular}{|l|c|c|c|c|}
\hline
\textbf{Method} & \textbf{Top-35 Acc} & \textbf{Top-5 Acc} & \textbf{Top-1 Acc} & \textbf{Speed} \\
\hline
YOLO-E + LightGlue & 0.6404 & 0.4858 & 0.4067 & 15s \\ \hline
PRISM  &  &  &  &  \\
(SigLIP + YOLO-E + LightGlue) & \textbf{0.7442} & \textbf{0.5142} & \textbf{0.4279} & \textbf{725ms} \\
\hline
\end{tabular}
}
\vspace{-0.2cm}
\caption{Comparison of retrieval accuracy and inference speed between our full model and the variant without SIGLIP.}
\label{tab:retrieval_results_syl}
\end{table}

As shown in Tab.~\ref{tab:retrieval_results_syl}, removing SigLIP results in a notable drop in retrieval accuracy across all metrics. The Top-35 accuracy decreases from 0.7442 to 0.6404, and Top-1 accuracy drops from 0.4279 to 0.4067. This indicates that YOLO-E's class-level filtering is too coarse to serve as an effective standalone retrieval stage, especially considering that product categories like ``canned" contain over 100 visually similar items. In such cases, LightGlue must compare the query against a much larger candidate pool, which not only increases inference time significantly (from 725ms to 15s) but also makes it more prone to incorrect matches.

In contrast, SigLIP reduces the candidate pool to the top 35 semantically similar products across all categories, enabling a much more efficient and focused matching process. These results highlight that semantic filtering with SigLIP is crucial for both improving retrieval accuracy and achieving real-time performance in retail environments.

\subsection{Importance of Stage 2 - Segmentation}
\begin{figure}[h]
    \centering
    \includegraphics[width=0.8\linewidth]{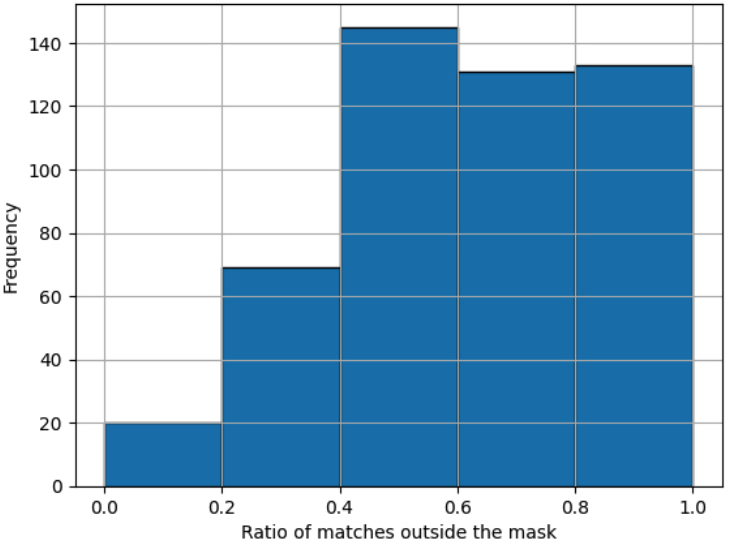}
    \vspace{-0.2cm}
    \caption{Histogram of the ratio of point matches that fall outside the segmentation mask to the total number of point matches.}
    \label{fig:segmentation-hist}
\end{figure}

To validate the effectiveness of segmentation, we analyze the spatial distribution of feature matches across query and gallery images by computing the ratio of matches that fall outside the product itself (out\_mask) to the total number of matches, defined as $\text{out\_mask} / (\text{out\_mask} + \text{in\_mask})$. Higher ratio indicates that most matches are not localized within the product region, suggesting that background features distract the matching algorithm.~Fig.~\ref{fig:segmentation-hist} shows the histogram of this ratio across the dataset, and demonstrates a high concentration of matches within the background region, confirming the importance of segmentation masks in improving matching quality.

\subsubsection{SAM vs. YOLO-E}
We also examine the impact of using a different segmentation model within our pipeline by replacing YOLO-E with the Segment Anything Model (SAM). As both methods are applied after the SigLIP stage, this comparison illustrates the effect of the segmentation step on final retrieval accuracy and speed.

Table~\ref{tab:ablation_sam} shows that YOLO-E outperforms SAM on Top-5 and Top-1 accuracy metrics.~More specifically, YOLO-E achieves 0.5142 Top-5 and 0.4279 Top-1 accuracy, whereas SAM scores 0.4979 and 0.3858, respectively. This suggests that YOLO-E provides more precise segmentation, which in turn improves downstream feature matching. In addition SAM is slightly slower (975ms vs. 725ms) than our pipeline that uses YOLO-E. These results confirm that YOLO-E offers a better performance considering both speed and retrieval accuracy.
\begin{table}[h!]
\centering
\resizebox{\columnwidth}{!}{%
\begin{tabular}{|l|c|c|c|}
\hline
\textbf{Segmentation Appr.} &  \textbf{Top-5 Acc} & \textbf{Top-1 Acc} & \textbf{Speed} \\

\hline
SAM & 0.4979 & 0.3858 & 975ms \\
YOLO-E & \textbf{0.5142} & \textbf{0.4279} & \textbf{725ms} \\
\hline
\end{tabular}
}
\vspace{-0.2cm}
\caption{Comparison of retrieval accuracy and inference speed when using SAM and YOLO-E for the segmentation step.}
\label{tab:ablation_sam}
\end{table}

\subsubsection{YOLOv8 vs. YOLO-E}
We also performed an experiment to compare YOLO-E with YOLOv8~\cite{reis2024yolov8}. 
To conduct this evaluation, we trained both models on 2,535 randomly selected images from our dataset and used 364 held-out test images to compute performance metrics. The results in Table~\ref{tab:yolo_comparison} show that YOLO-E consistently outperforms YOLOv8 in both the precision of segmentation and retrieval accuracy. This improvement stems from YOLO-E’s ability to generalize better to novel product classes with limited or no training examples, thanks to its open-vocabulary design. Additionally, YOLO-E produces more accurate segmentation masks. Thus, we have chosen YOLO-E as the segmentation approach in stage 2 of our framework for improved visual matching and product retrieval performance. 
\begin{table}[h!]
\centering
\begin{tabular}{|l|c|c|c|}
\hline
\textbf{Model} & \textbf{Box (mAP)} & \textbf{Recall (r)} & \textbf{mAP@[.50:.95]} \\
\hline
YOLOv8 & 0.988 & 0.975 & 0.801 \\
YOLO-E & \textbf{0.991} & \textbf{0.984} & \textbf{0.905} \\
\hline
\end{tabular}
\vspace{-0.2cm}
\caption{Comparison of YOLOv8 and YOLO-E performance.}
\label{tab:yolo_comparison}
\end{table}

\subsection{Importance of Stage 3 - Pixel-wise Matching}
In the context of pixel-wise matching, we compare LoFTR with LightGlue.~Similar to above, this step is applied after SigLIP provides the top 35 returns. As shown in Table~\ref{tab:retrieval_results_lfm}, LightGlue achieves higher Top-5 (0.5142) and Top-1 (0.4279) accuracy compared to LoFTR (0.4982 Top-5 and 0.4001 Top-1), demonstrating better fine-grained matching performance. In addition to its accuracy gains, LightGlue is substantially faster, with an average inference time of 725ms per query, whereas LoFTR requires 2.9 seconds. This highlights LightGlue’s superior efficiency–accuracy trade-off and supports its use for real-time product retrieval in retail environments.
\begin{table}[h!]
\centering
\resizebox{0.8\columnwidth}{!}{%
\begin{tabular}{|l|c|c|c|}
\hline
\textbf{Method} &  \textbf{Top-5 Acc} & \textbf{Top-1 Acc} & \textbf{Speed} \\

\hline
LoFTR & 0.4982 & 0.4001 & 2.9s \\
LightGlue & \textbf{0.5142} & \textbf{0.4279} & \textbf{725ms} \\
\hline

\end{tabular}
}
\vspace{-0.2cm}
\caption{Comparison of retrieval accuracy and inference speed between LightGlue and LoFTR.}
\label{tab:retrieval_results_lfm}
\end{table}

\subsection{CLIP vs. YOLO-E + CLIP}
Another intuitive approach is to combine YOLO-E with CLIP by first segmenting the product with YOLO-E and then applying CLIP feature extraction to the cropped image. However, our experiments, summarized in Table~\ref{tab:clip_results} show that this does not lead to significant improvements compared to using CLIP alone. This suggests that simply cropping the image to eliminate background clutter is not enough to overcome CLIP’s limitations in fine-grained product differentiation.

One possible explanation is that CLIP embeddings are already robust to background noise due to their contrastive training with diverse natural image-text pairs. Cropping the image might even remove contextual clues that CLIP relies on for semantic understanding.
\begin{table}[h!]
\centering
\resizebox{\columnwidth}{!}{%
\begin{tabular}{|l|c|c|c|}
\hline
\textbf{Method} & \textbf{Top-35 Acc} & \textbf{Top-5 Acc} & \textbf{Top-1 Acc} \\
\hline
CLIP & \textbf{0.5300} & \textbf{0.3137} & 0.1704 \\
YOLO-E + CLIP & 0.5100 & 0.2792 & \textbf{0.1754} \\
\hline
\end{tabular}
}
\vspace{-0.25cm}
\caption{Comparison of retrieval accuracy (Top-35, Top-5, and Top-1) between standard CLIP and YOLO-E followed by CLIP.}
\label{tab:clip_results}
\vspace{-0.1cm}
\end{table}

\section{Conclusion}
\label{sec:conclusion}

We have proposed an efficient and effective framework for product retrieval in retail settings using camera images from shopping carts.~Our proposed PRISM addresses the key challenge of distinguishing visually similar items, such as canned tomato paste versus canned fire roasted tomatoes, by an hybrid approach combining semantic retrieval and pixel-level feature matching while decreasing the computational demand. PRISM leverages the advantages of both vision-language model-based and pixel-wise matching approaches, and employs a three-step framework. 
Through comprehensive experiments, we have shown that our proposed PRISM provides the best Top-1 accuracy, outperforming recent vision-language models, including JinaCLIP, LongCLIP and SigLIP. While models like SigLIP perform well in broader Top-35 retrieval metrics, our PRISM enables more accurate discrimination between products with high inter-class similarity by focusing on subtle visual cues often missed by global models, and offers a more effective trade-off between fine-grained precision and runtime efficiency. We have demonstrated the critical role of semantics-based filtering and segmentation in real-world scenarios, where product images frequently include distracting backgrounds and visually complex environments. We have also performed ablation studies showing the role of different stages of PRISM on the overall performance. 

{
    \small
    \bibliographystyle{ieeenat_fullname}
    \bibliography{main}
}

\end{document}